\documentclass{article}

\PassOptionsToPackage{numbers}{natbib}
\usepackage[final]{nips_2016}
\usepackage{natbib}

\usepackage[utf8]{inputenc} 
\usepackage[T1]{fontenc}    
\usepackage{hyperref}       
\usepackage{url}            
\usepackage{booktabs}       
\usepackage{amsfonts}       
\usepackage{nicefrac}       
\usepackage{microtype}      

\usepackage{graphics} 
\usepackage{epsfig} 
\usepackage{amsmath} 
\usepackage{amssymb}  
\usepackage{algorithm}
\usepackage{algpseudocode}
\usepackage{mathtools}
\usepackage{graphicx,subfigure}
\usepackage[usenames,dvipsnames]{color}

\graphicspath {{figures/}}

\title{Multiscale Inverse Reinforcement Learning using Diffusion Wavelets}

\author{
  Jung-Su Ha and  Han-Lim Choi\\
  Dept. of Aerospace Engineering, KAIST\\
  \texttt{\{wjdtn1404, hanlimc\}@kaist.ac.kr} \\
}

\begin{document}

\maketitle

\begin{abstract}
This work presents a multiscale framework to solve an inverse reinforcement learning (IRL) problem for continuous-time/state stochastic systems.
We take advantage of a diffusion wavelet representation of the associated Markov chain to abstract the state space. This not only allows for effectively handling the large (and geometrically complex) decision space but also provides more interpretable representations of the demonstrated state trajectories and also of the resulting policy of IRL.
In the proposed framework, the problem is divided into the global and local IRL, where the global approximation of the optimal value functions are obtained using coarse features and the local details are quantified using fine local features. An illustrative numerical example on robot path control in a complex environment is presented to verify the proposed method.
\end{abstract}

\section{Introduction}
In this paper, we address an inverse reinforcement learning (IRL) problem (or often equivalently called an inverse optimal control problem) for robots operated in a complex environment over a long time horizon, where its forward problem is a continuous time/continuous stochastic optimal control problem called linearly solvable optimal control (LSOC).
The objective of an IRL problem is to recover the value and cost functions as well as the optimal policy when experts' demonstrations are given.
A general method to solve IRL problem for standard MDP often involves the procedure of solving the corresponding forward problem in every iteration~\cite{ziebart2008maximum}, while a method for LSOC does not~\cite{dvijotham2010inverse}.
The IRL solution method for LSOC is formulated as a convex optimization problem, where its gradient and Hessian are obtained analytically.
However, the optimization tends to be intractable as the size of the problem increases and moreover, in real situation, a demonstration data set may be not sufficient to represent whole state space.
Finding a sparse structure of the problem and representing the problem with few meaningful bases are essential for obtaining the solution efficiently.

To address a large-scale problems effectively, it is conceivable that the hallmark of human intelligence could provide some insights - in particular, this work notes \textit{multiscale} and \textit{hierarchical} structure of human decision making.
Suppose that someone currently writing a paper at his/her office desk wants to get out of the building; the office is located in the third floor of the building and there is one set of staircases and an elevator.
Then, what would this person's control policy look like?
This person would not try to figure out what he/she should do for all possible situations he/she could face like the standard value function-based approach; instead, he/she would figure out which building gate he/she would use, whether to take the elevator or the stairs, which door he would exit from the room (if there are more than one), etc. A detailed plans such as ``which particular start he/she should put their left foot,'' would be determined later in the process of executing a piece of overall plan, for example, ``go downstairs using the staircase.''
It should be noted that this human-like decision takes advantage of the underlying (multiscale) hierarchical structure of state space; in the above example, detailed aspects such a particular certain sequence of stairs to put on is abstracted by just a single notion of staircase.

\begin{figure}[t]
	\centering
	\includegraphics*[width=.7\columnwidth, viewport=40 85 920 515]{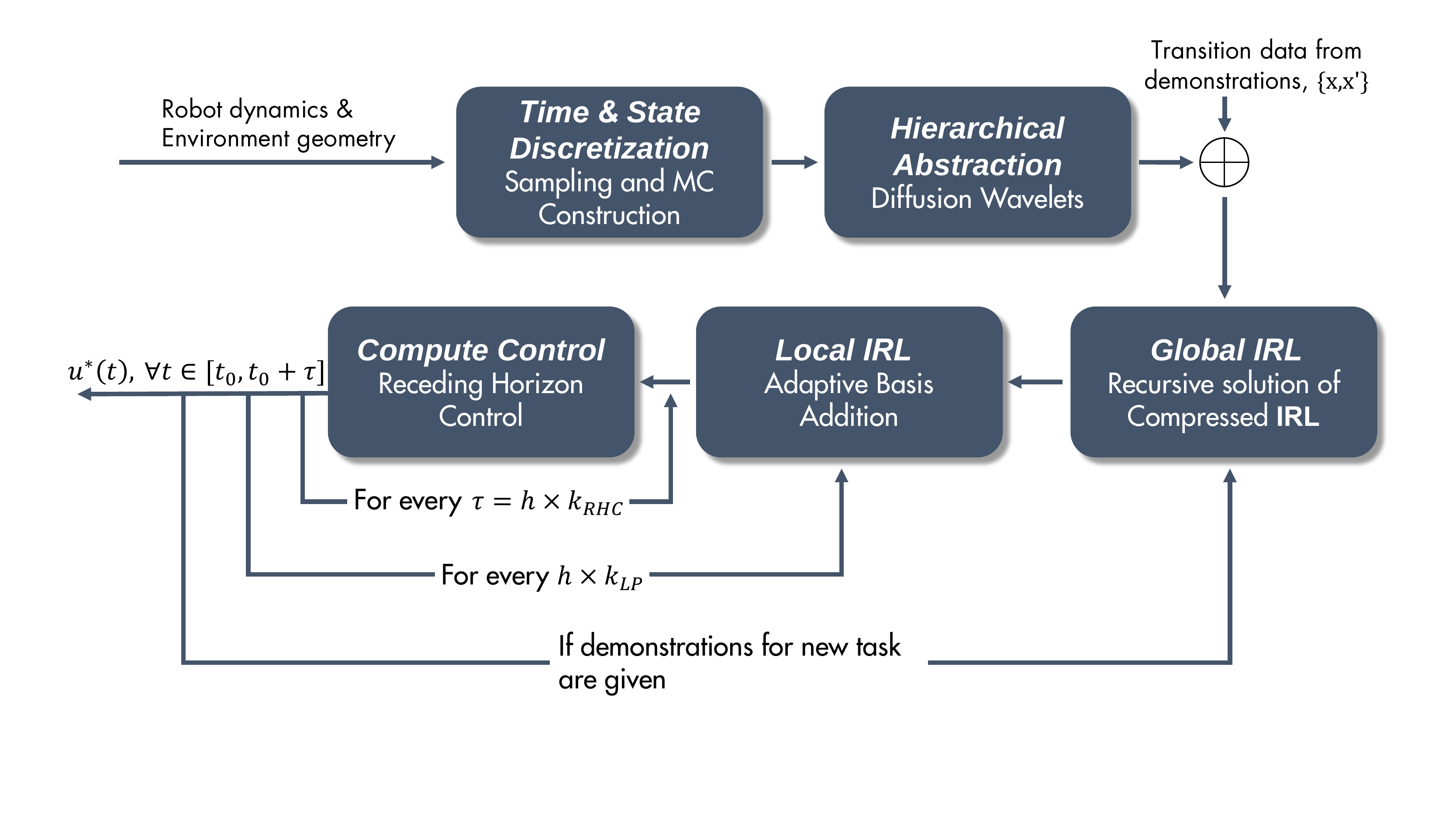}
	\caption{Proposed framework}
	\label{fig:framework}
\end{figure}
Same intuition can be applied to the inverse inference problem.
By utilizing it, the key contribution of this paper is to present a systematic framework for solving IRL, as depicted in Fig. \ref{fig:framework}. The framework consists of five phases: (i)  In the discretization phase, Markov chain associated to the robot dynamics is constructed by sampling a set finite set of states in state-space. (ii) In the abstraction phase, the hierarchical bases structure is obtained using the diffusion wavelet method. (iii) In global IRL phase, an IRL problem constructed only using the coarse bases (or on ``abstract-state'') is solved; this IRL is much more tractable to handle than using the original bases set. (iv) In the local planning phase, the finer bases located in focused regions where the demonstration data visit frequently are sought for and detailed solution associated with these focused regions are computed. (vi) In the control phase, a continuous control sequence is computed and applied to the robot in a receding horizon fashion. Rest of the paper is primarily focused on elaborating the details of this framework, followed by numerical example for validation of the method.

\section{Inverse Reinforcement Learning}
\subsection{Linearly-solvable optimal control}
Consider a stochastic dynamics whose deterministic drift term is affine in control input:
\begin{equation}
d\mathbf{x} =  \mathbf{f}(\mathbf{x})dt+ G(\mathbf{x})(\mathbf{u}dt + \sigma dw) \label{eq:conti_dyn}
\end{equation}
where $\mathbf{x}\in\mathcal{X}\subset\mathbb{R}^{d_x}$, $\mathbf{u}\in\mathbb{R}^{d_u}$ and $w$ are a state, control vector and an $d_u$-dimensional Brownian motion process, respectively.
Let functions $q: \mathcal{X} \rightarrow \mathbb{R}$ and $\pi_c: \mathcal{X}\rightarrow\mathbb{R}^{d_u}$ be an instantaneous state cost rate and a control policy, respectively and define an instantaneous cost rate as $l(\mathbf{x},\mathbf{u}):=q(\mathbf{x})+\frac{1}{2\sigma^2}\mathbf{u}^T\mathbf{u}$.
Then the cost functional is given as:
\begin{equation}
J_{conti}^{\pi_c}(\mathbf{x}) = \lim_{t_f\rightarrow\infty}\frac{1}{t_f}E\left[\int^{t_f}_0 l(\mathbf{x}(t), \pi_c(\mathbf{x}(t))dt\right]. \label{eq:conti_cost}
\end{equation}
The problem with the cost function (\ref{eq:conti_cost}) and dynamics (\ref{eq:conti_dyn}) is called the infinite horizon average cost stochastic optimal control (SOC) problem or is referred as linearly-solvable optimal control (LSOC) problem since its solution is obtained from the linear partial differential equation \cite{todorov2009efficient}.

If the time-axis is discretized by a time step $h$, the transition probability of one step without/with any control input is defined as:
\begin{equation}
\mathbf{x}[k+1] \sim p(\cdot|\mathbf{x}[k]),~\mathbf{x}[k+1] \sim \pi(\cdot|\mathbf{x}[k]), \label{eq:discrete_dyn}
\end{equation}
which are called the passive and controlled dynamics, respectively.
The passive and controlled dynamics are approximated as $\mathcal{N}(\mathbf{y}; \mu(h), \Sigma(h))$, where $\mathcal{N}$ is a Gaussian distribution with a mean $\mu(h)$ and covariance $\Sigma(h)$.
For small $h$, the Kullback-Leibler divergence between two distributions is approximated as $D_{KL}(\pi(\cdot|\mathbf{x})||p(\cdot|\mathbf{x})) = \frac{h}{2\sigma^2}\mathbf{u}'\mathbf{u}$.
Therefore, the cost functional (\ref{eq:conti_cost}) is written in the discrete time setting:
\begin{align}
J^\pi(\mathbf{x}) = \lim_{K\rightarrow\infty}\frac{1}{K}E\left[\sum^{K}_{k=0} hq(\mathbf{x}[k])+D_{KL}\left(\pi(\cdot|\mathbf{x}[k])||p(\cdot|\mathbf{x}[k])\right)\right]. \label{eq:discrete_cost}
\end{align}
Moreover, the state space can be discretized by sampling a set of states $X=\{\mathbf{x}_n\}$~\citep{ha2016multi}.
Transition probability matrix for passive dynamics $P$, where $P_{nm}$ means a transition probability from $\mathbf{x}_n$ to $\mathbf{x}_m$, is approximated via Gaussian distribution as:
\begin{equation}
P_{nm} = \frac{\mathcal{N}(\mathbf{x}_m:\mu(h), \Sigma(h))}{\sum_{m'}\mathcal{N}(\mathbf{x}_{m'}:\mu(h), \Sigma(h)}, \label{eq:approx_Gaussian}
\end{equation}
where $\mu(h)$ and $\Sigma(h)$ are computed by integrating moment dynamics of linearized SDE for $t\in[0,h]$:
\begin{align}
\dot{\mu}(t) = A\mu(t)+\mathbf{c},~\dot{\Sigma}(t) = A\Sigma(t)+\Sigma(t)A'+BB' \label{eq:Cov_dyn}
\end{align}
from $\mu(0)=\mathbf{x}_n,~\Sigma(0) = 0$, where $A = \left.\frac{d\mathbf{f}}{d\mathbf{x}}\right|_{\mathbf{x}=\mathbf{x}_n}$, $B=\sigma G(\mathbf{x}_n)$ and $\mathbf{c}=\mathbf{f}(\mathbf{x}_n)-A\mathbf{x}_n$.
One can truncate tails of Gaussian distribution to make $P$ sparse.
With a set of discrete states, $X$, the state-space as well as time-axis discretized version of SOC is formulated as the Markov decision process (MDP) of which cost function is given by (\ref{eq:discrete_cost}).
This type of MDP is called the linearly-solvable MDP \citep{todorov2009efficient} and its solution is known to converge to SOC solution as $|X|\rightarrow\infty$ and $h\rightarrow0$.

Define the optimal cost-to-go value $c := \min_\pi J^\pi(\mathbf{x})$, the differential cost-to-go function $v(\mathbf{x})$, the (differential) desirability function $z(\mathbf{x}) = \exp(-v(\mathbf{x})),$ and the linear operator $\mathcal{G}[z](\mathbf{x})=\sum_{\mathbf{x}'}p(\mathbf{x}'|\mathbf{x})z(\mathbf{x}')$.
Then $z(\mathbf{x})$ satisfies the following linear Bellman equation:
\begin{equation}
\exp(-c)z(\mathbf{x})=\exp(-hq(\mathbf{x}))\mathcal{G}[z](\mathbf{x}), \label{eq:lin_Bellman}
\end{equation}
and the optimal policy is obtained analytically:
\begin{equation}
\pi^*(\mathbf{x}'|\mathbf{x}) = \frac{p(\mathbf{x}'|\mathbf{x})z(\mathbf{x}')}{\mathcal{G}[z](\mathbf{x})}. \label{eq:opt_pi}
\end{equation}
For more details of problem formulation and discrete approximation method, we would refer the reader to \citep{ha2016multi} and references therein.

\subsection{Inverse reinforcement learning for LSOC problem}
While the objective of (forward) SOC problem is to find the optimal control policy for the given system and cost function, the objective of inverse reinforcement learning (IRL) problem is to recover the value and cost functions as well as the optimal policy when experts' demonstrations are given \cite{dvijotham2010inverse,ziebart2008maximum}.
Suppose a dataset of transitions $\{\mathbf{x}_n,\mathbf{x}'_{n}\}_{n=1,\cdots,N}$ is obtained from the optimal policy \eqref{eq:opt_pi}.
Let $\mathbf{v}$ be a vector representation of value function $v(\cdot)$.
Then, the negative log-likelihood of the dataset is given as:
\begin{equation}
L[\mathbf{v}] = \mathbf{a}^T\mathbf{v} + \mathbf{b}^T\log(P\exp(-\mathbf{v})), \label{eq:likelihood}
\end{equation}
where each component of $\mathbf{a}$ and $\mathbf{b}$ represent visitation counts of $\mathbf{x}_n'$ and $\mathbf{x}_n$, respectively.
Since $L$ is convex and its gradient and Hessian are computed analytically, it can be minimized by applying iterative second order convex optimization methods.
Once the value function $\mathbf{v}$ is obtained, the cost function $q(\cdot)$ and the optimal policy $\pi^*(\cdot|\cdot)$ can be recovered directly from \eqref{eq:lin_Bellman} and \eqref{eq:opt_pi}, respectively.
In real situation, however, $\mathbf{a}$ and $\mathbf{b}$ are sparse since the dataset is not sufficient, so it is impossible to compute $v(\cdot)$ over whole state space.
Also, the optimization procedure in \eqref{eq:likelihood} gets intractable for problems having the large size of the state space.
To represent the problem efficiently, a linear value function approximation can be used:
\begin{equation}
\hat{\mathbf{v}} = \Phi\mathbf{w},
\end{equation}
where each column of $\Phi$ represents a feature (or basis) and $\mathbf{w}$ is weight.
Note that $L[\mathbf{w}]$ is also convex.
In this work, we obtain the hierarchical structure of feature sets which is naturally induced from the passive (diffusion) dynamics of the system and utilize those set of features to solve IRL problem efficiently.

\section{Multiscale Inverse Reinforcement Learning}
\subsection{Multiscale feature extraction: Diffusion Wavelets}
From now on, we consider $T=P'$ for notion simplicity; then $T_{nm}$ represents a transition probability from $\mathbf{x}_m$ to $\mathbf{x}_n$.
The Markov chain, $T$, obtained by discretizing a diffusion process ((\ref{eq:conti_dyn}) with $\mathbf{u}=0$) is known to have some interesting properties: \textit{local}, \textit{smoothing} and \textit{contractive} \citep{coifman2006diffusion}.
From any initial point, $\delta_m$, the state (numerically) transitions to only a few its neighbors (i.e., $T\delta_m$ has a small support) and $T^j\delta_m$ is a smooth probability distribution.
Also since $||T||_2\leq1$, a dimension of a subspace, $V_j$, which is $\epsilon$-spanned by $\{T^j\delta_m\}_{\mathbf{x}_m\in X}$ monotonically decreases as $j$ increases and $V_0\supseteq V_1\supseteq\cdots\supseteq V_j\supseteq\cdots$; especially for an irreducible Markov chain, $\text{dim}(V_j)\rightarrow 1$ as $j$ increases and a limit of $V_j$ corresponds to the stationary distribution of the Markov chain.

Let $W_j$ be an orthogonal complement of $V_{j+1}$ into $V_j$, i.e., $V_j = V_{j+1}\oplus W_j$ and suppose the orthonormal bases $\Phi_j$ and $\Psi_j$ span $V_j$ and $W_j$, respectively.
By using aforementioned properties of $T$, \textit{Diffusion wavelets} constructs a hierarchical structure of a set of well-localized bases $\Phi_j$ and $\Psi_j$ called \textit{scaling} and \textit{wavelet functions}, respectively, in order that the subspace spanned by feature set $[\Phi_j]_{\Phi_0}$ is $j\epsilon$-close to the subspace spanned by $\{T^{1+2+2^2+\cdots+2^{j-1}}\delta_m=T^{2^j-1}\delta_m\}_{\mathbf{x}_m\in X}$.
Roughly speaking, $\Phi_j$ and $\Psi_j$ represent smooth bump function and oscillatory function, respectively.
We omit the procedure of Diffusion wavelets algorithm because of the space limitation and would refer the readers to \citep{coifman2006diffusion,ha2016multi} for more details.

Let $[B]_{\Phi_j}$ be a set of vectors $B$ represented on a basis $\Phi_j$, where the columns of $[B]_{\Phi_j}$ are the coordinates of the vectors $B$ in the coordinates $\Phi_j$.
A set of features at level $j$ can be written in the original coordinate (or can be \textit{unpacked}) as:
\begin{align}
\Phi_j=[\Phi_j]_{\Phi_0}=[\Phi_{j-1}]_{\Phi_0}[\Phi_{j}]_{\Phi_{j-1}}=[\Phi_1]_{\Phi_0}\cdots[\Phi_{j-1}]_{\Phi_{j-2}}[\Phi_{j}]_{\Phi_{j-1}},
\end{align}
which is represented as a $|X|\times|X_j|$ matrix.
Note that each column of $[\Phi_j]_{\Phi_0}$ can be viewed as an ``abstract-state" of the original Markov chain.
At the scale $j$, there are only $|X_j|$ meaningful combinations of states and each combination, $[\Phi_j]_{\Phi_0}$, represents ``abstract-state''.

\subsection{IRL with hierarchical multiscale feature sets}
Rather than solving original $|X|$-dimensional optimization problem, we can treat the lower-dimensional coarsened problem.
Suppose a set of ``abstract-state" at level $j$, $\Phi_j$, is utilized as a set of features, which means the problem is viewed in a lower resolution with $2^j$ time scale.
Then, $\mathbf{v}$ is approximated as a linear combination of this feature set as $\hat{\mathbf{v}}_j=\Phi_j\mathbf{w}_j$, and the optimization problem \eqref{eq:likelihood} is also written as:
\begin{equation}
L[\mathbf{w}_j] = \mathbf{a}^T\Phi_j\mathbf{w}_j + \mathbf{b}^T\log(P\exp(-\Phi_j\mathbf{w}_j)). \label{eq:likelihood_approx}
\end{equation}
The compressed problem \eqref{eq:likelihood_approx} is much more tractable than the original problem \eqref{eq:likelihood} if $|X_j|<<|X|$.
Note that due to its localization property, the features are naturally \textit{interpretable} (see Fig. \ref{fig:results} (a)--(d));
thus user can choose the appropriate level, considering trade-off between the size of the problem and the solution quality.
Also, the hierarchical structure of diffusion wavelet tree can be utilized to solve the problem more efficiently;
the solution of $j$th level, $\mathbf{w}_{j}$, can provide an \textit{initial guess} to $(j-1)$th level problem;
that is, the optimization at $(j-1)$th level starts from $\tilde{\mathbf{w}}_{j-1}= [\Phi_{j}]_{\Phi_{j-1}}\mathbf{w}_{j}$ by unpacking the $j$th level solution.
If $\hat{\mathbf{v}}_{j}$ is not sharply changed through the scale $j$, this initial guess would be near the optimum of $(j-1)$th level problem and the optimization procedure would rapidly converge to the minimum.

Suppose we solve $l$th level problem and have the approximate solution $\mathbf{v}_l$ by using features $\Phi_l$.
Then, we can achieve more exact solution by considering additional features from wavelet functions, $\Psi_{1:(l-1)}$ which are orthogonal to $\Phi_l$ (note that $V_0=V_l\oplus W_{l-1}\oplus W_{l-2}\oplus\cdots\oplus W_0$).
The wavelet bases are also built as being well-localized.
In this work, we utilize the intuition that the \textit{important} region where optimal policy frequently visits should have highest resolution \cite{ha2016multi,dvijotham2010inverse}.
The wavelet functions are evaluated as:
\begin{equation}
\mathbf{s} = \mathbf{b}^T|\Psi_{1:(l-1)}|,
\end{equation}
where the score $\mathbf{s}\in\mathbb{R}^{|X|-|X_l|}$ represents how each feature overlap with visitation counts of $\mathbf{x}'_n$.
By adding features with high scores and solving the corresponding problem, the value, cost functions and policy will have higher resolution in \textit{important} region.
Also, user can easily choose the additional features for their objective since the wavelet features are also \textit{interpretable}.

Finally, the continuous-time optimal policy can be extracted from the multiscale quantification of the optimal value structure for the interval $\tau=h\times k_{RHC}$ in a receding horizon control fashion.
The control can be computed by matching the $1^\text{st}$ order moment of original SDE (\ref{eq:conti_dyn}) and the optimal policy (\ref{eq:opt_pi}) for the MDP:
\begin{equation}
\mathbf{u}^*(t)=-\sigma^2G^Te^{A^T(\tau-t)}\Sigma(\tau)^{-1}(\mu(\tau)-\mathbf{y}_{new}), \label{eq:opt_con}
\end{equation}
where $\mathbf{y}_{new}= \sum_{\mathbf{y}\in X}\mathbf{y}Pr(\mathbf{x}[k_{RHC}]=\mathbf{y}|\mathbf{x}[0]=\mathbf{x}_{cur},\hat{\pi}^*)$ denotes the expected state after $\tau$ when the system follows the (approximate) optimal policy (\ref{eq:opt_pi}) from $\mathbf{x}_{cur}$.

\section{Numerical Example}
\begin{figure}[tp]
	\centering
	\subfigure[]{
		\includegraphics*[width=.23\columnwidth, viewport =52 30 380 300]{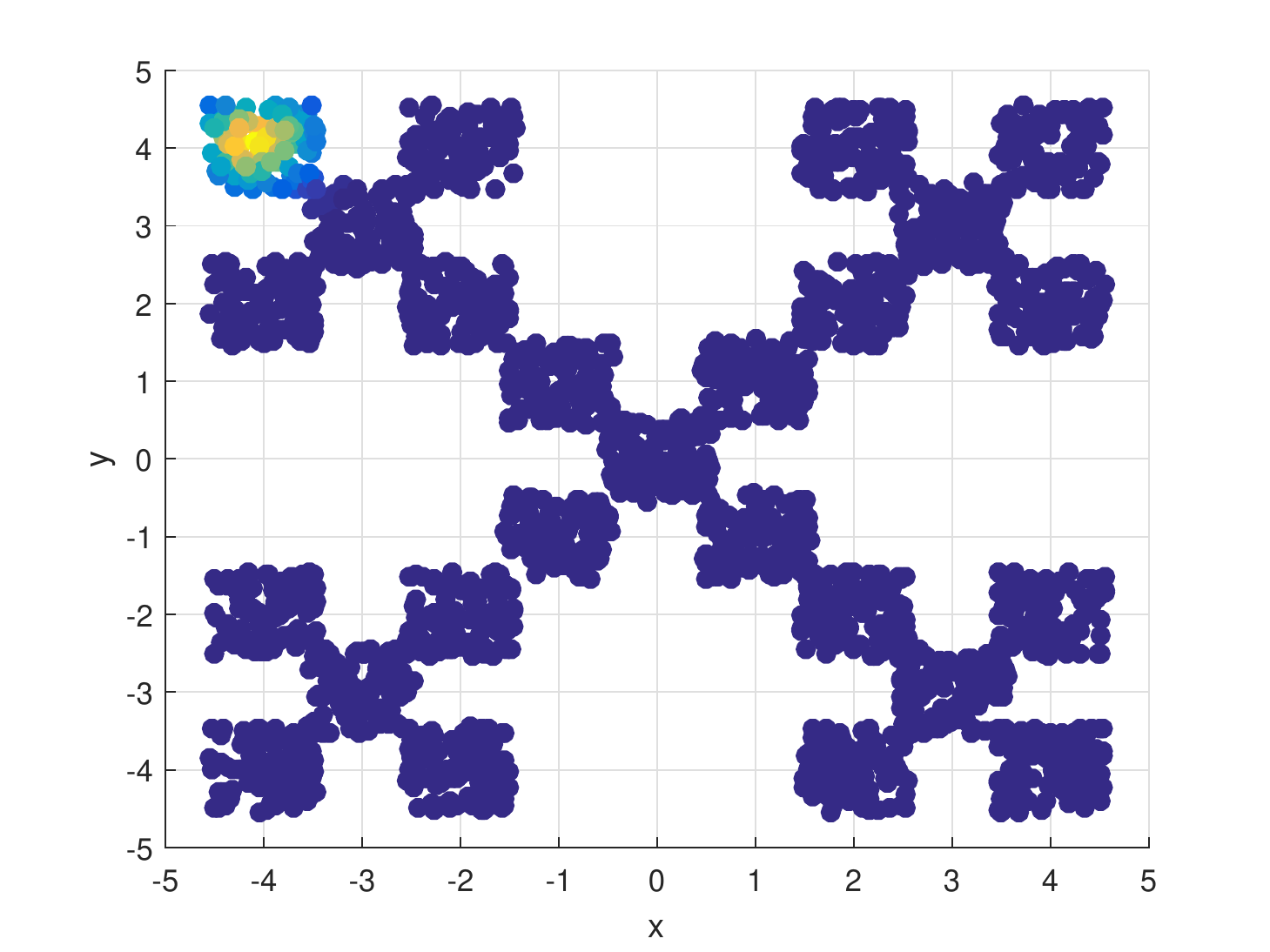}}
	\subfigure[]{
		\includegraphics*[width=.23\columnwidth, viewport =52 30 380 300]{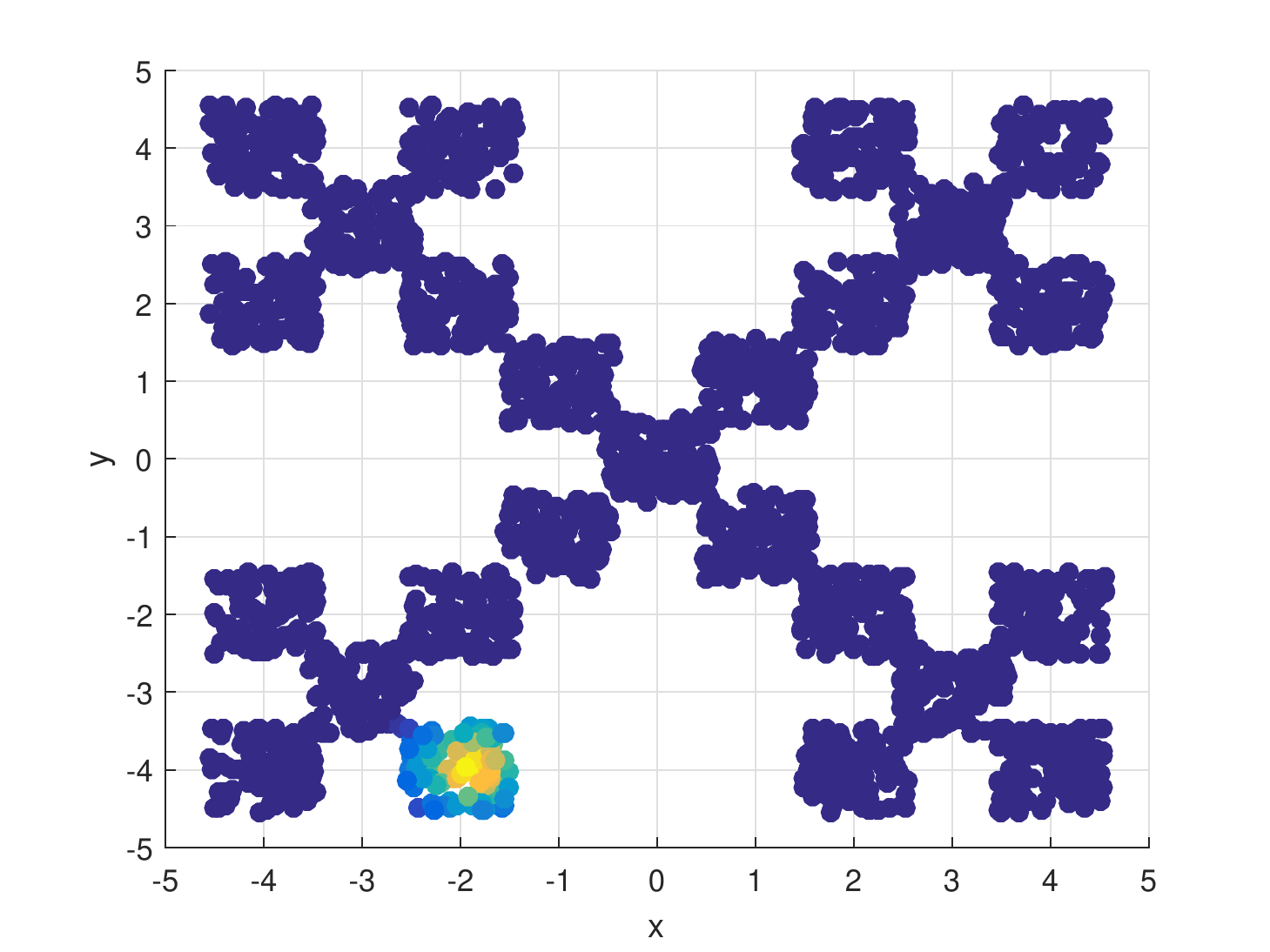}}
	\subfigure[]{
		\includegraphics*[width=.23\columnwidth, viewport =52 30 380 300]{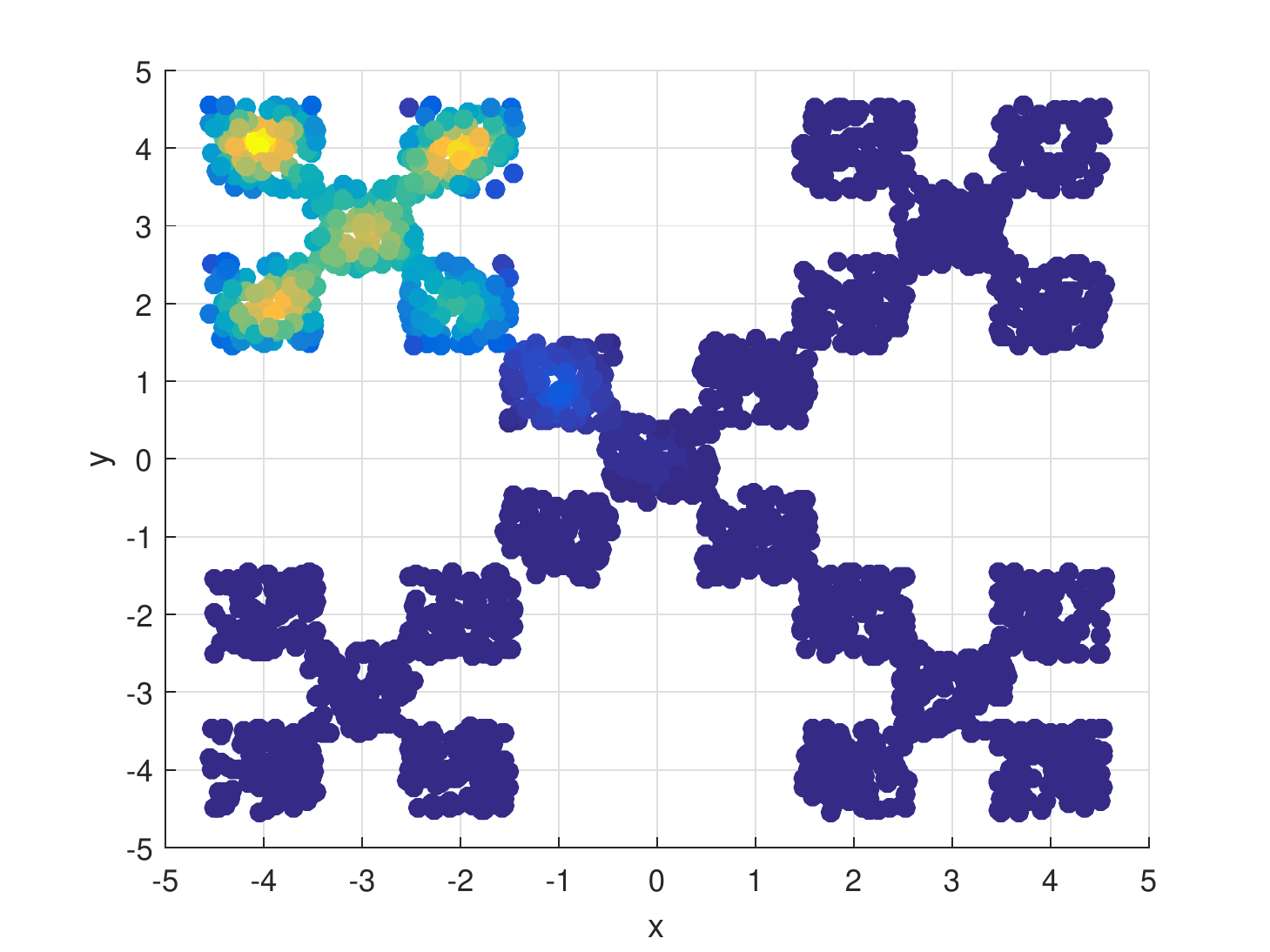}}
	\subfigure[]{
		\includegraphics*[width=.23\columnwidth, viewport =52 30 380 300]{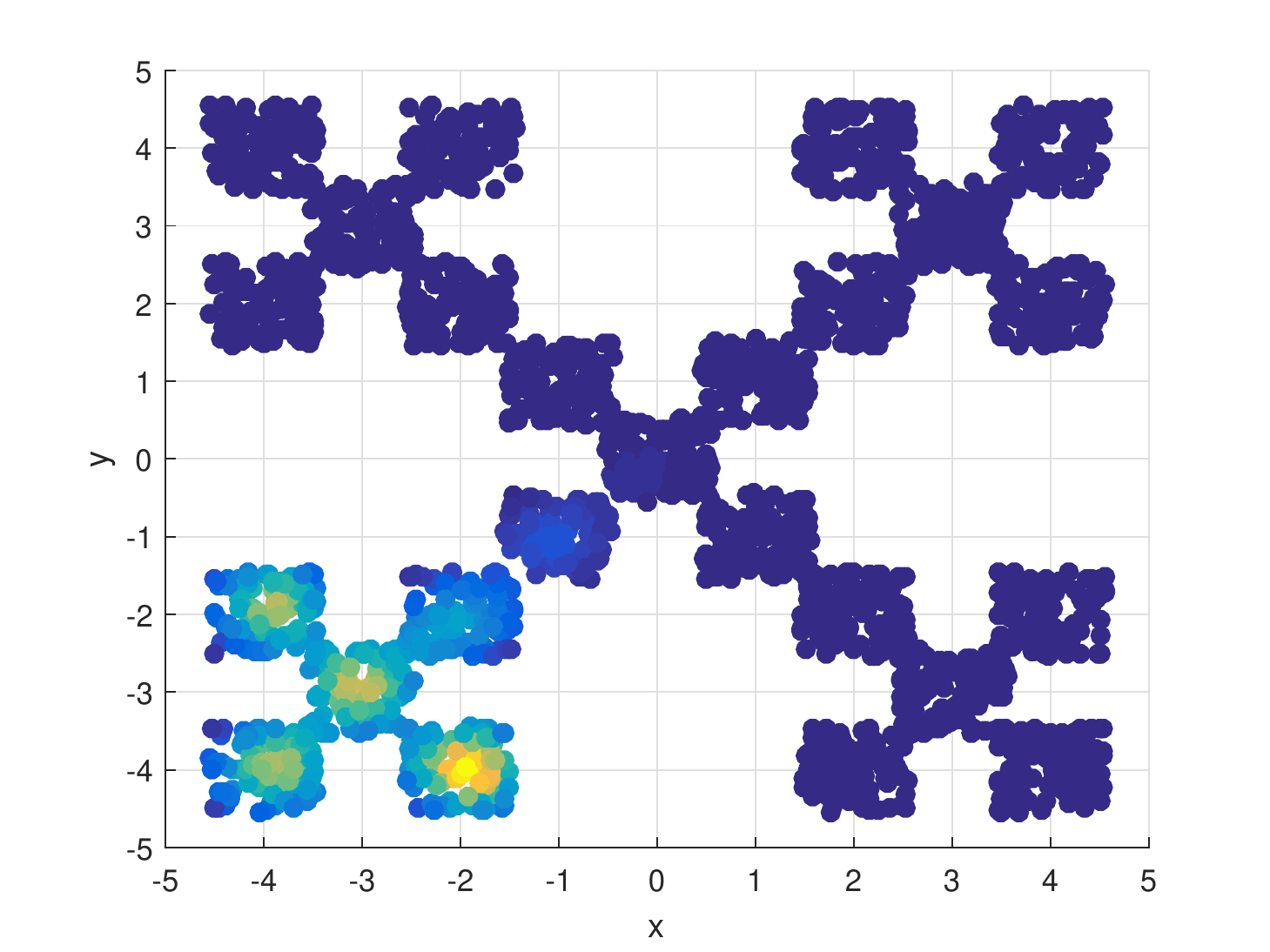}}
	\subfigure[$q(\mathbf{x})$]{
		\includegraphics*[width=.23\columnwidth, viewport =48 30 385 300]{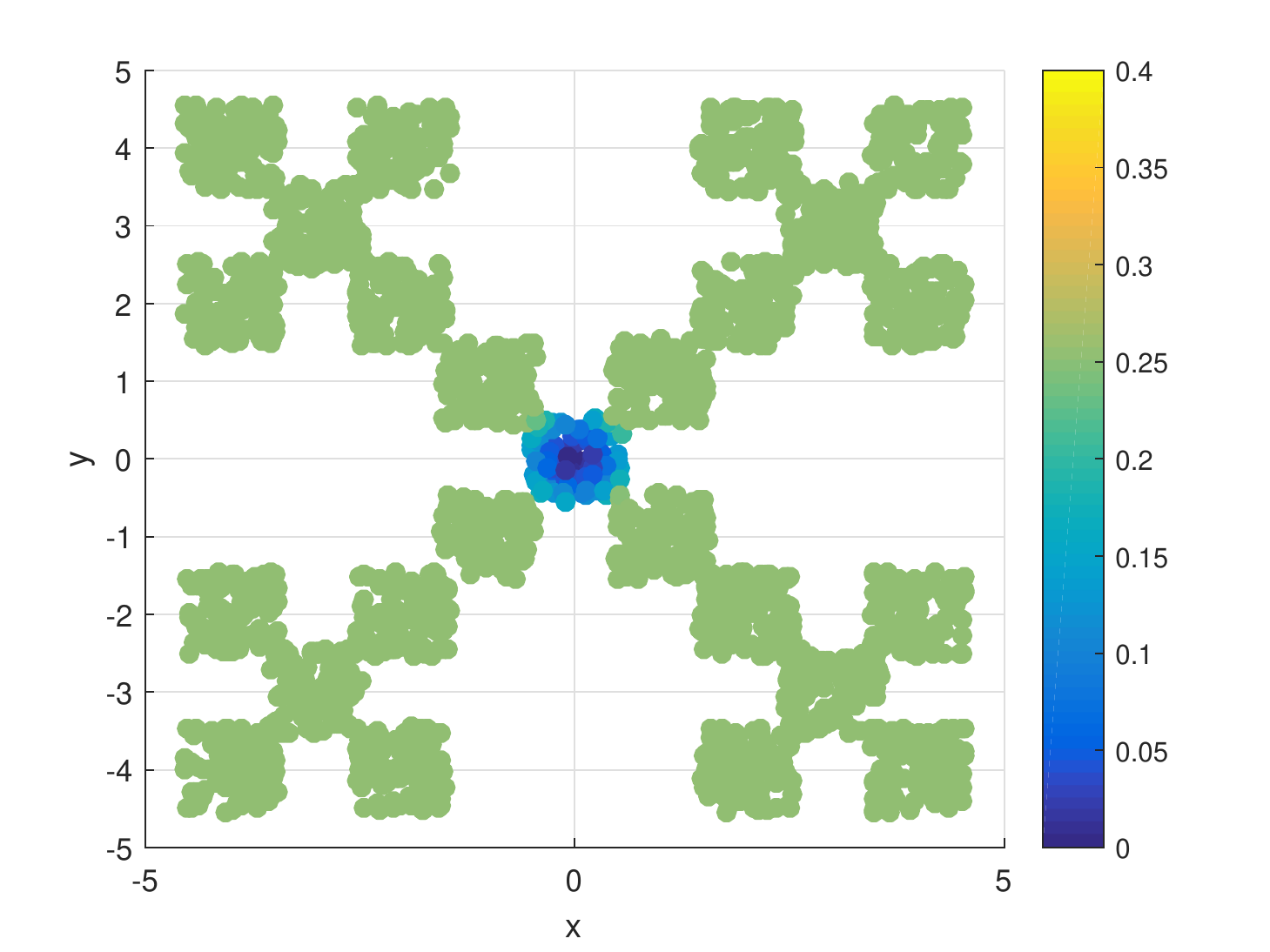}}
	\subfigure[$v(\mathbf{x})$]{
		\includegraphics*[width=.23\columnwidth, viewport =48 30 385 300]{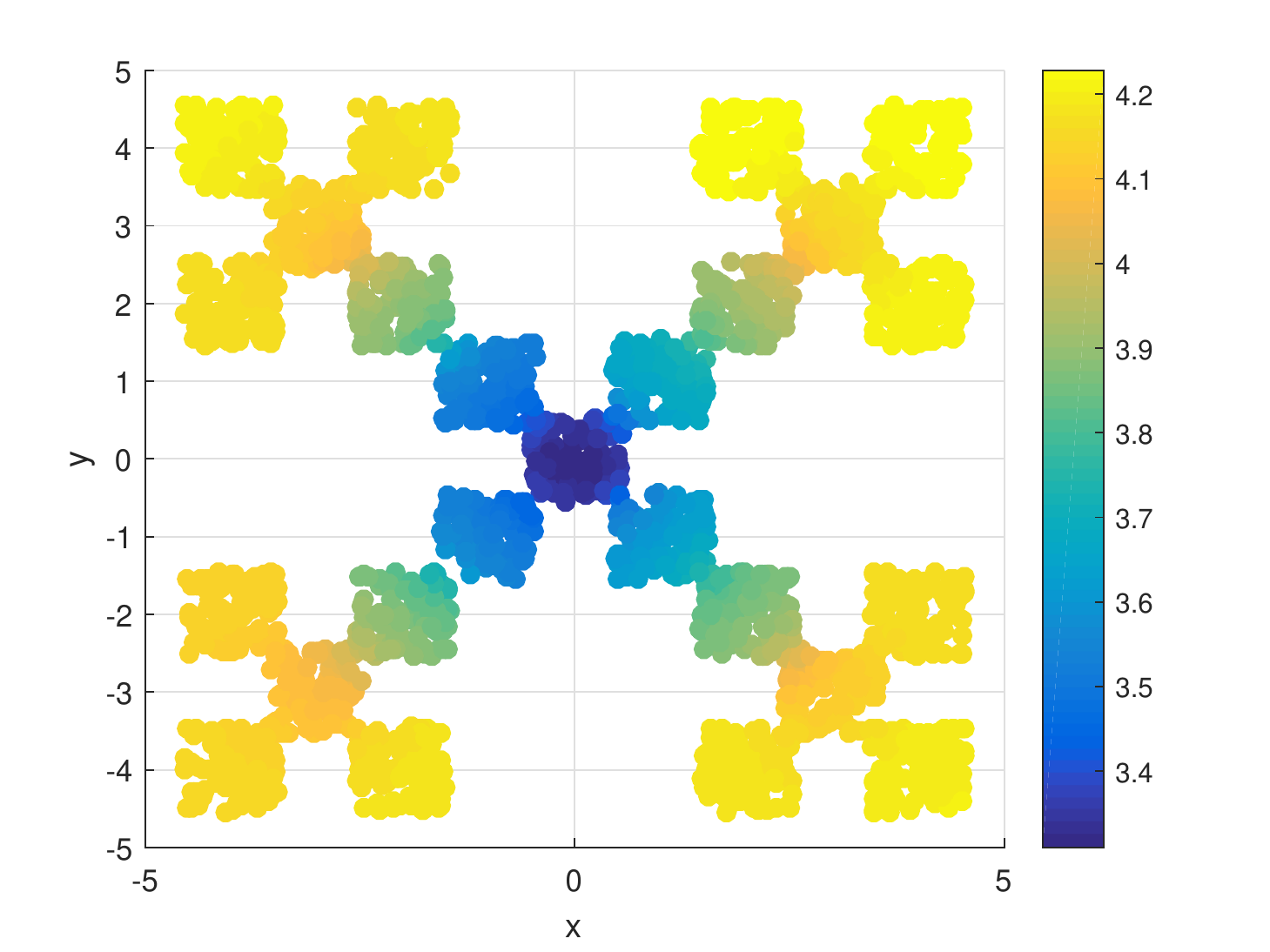}}
	\subfigure[$\hat{\mathbf{v}}_3$]{
		\includegraphics*[width=.23\columnwidth, viewport =48 30 385 300]{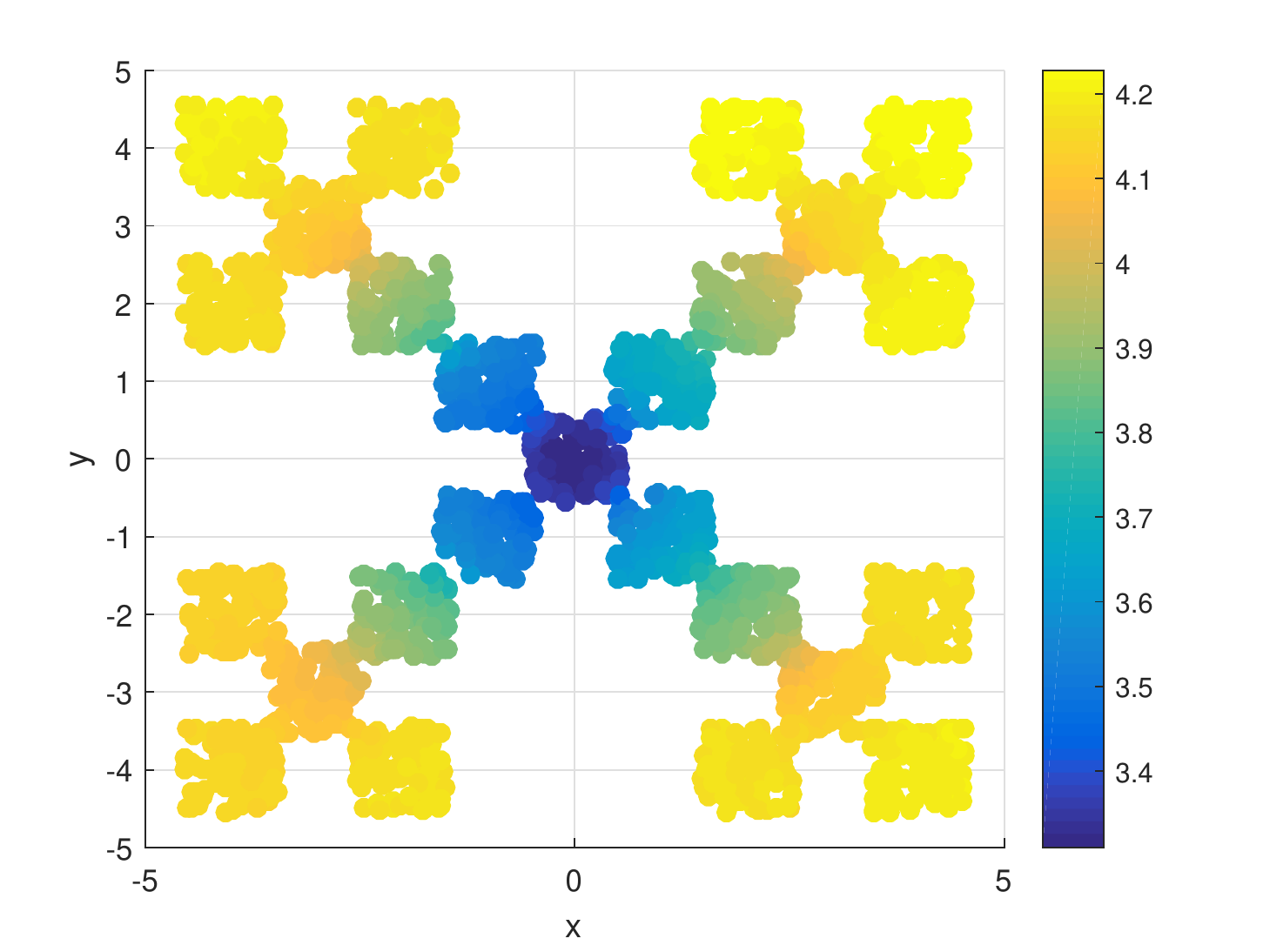}}
	\subfigure[$\hat{\mathbf{v}}_{8}$]{
		\includegraphics*[width=.23\columnwidth, viewport =48 30 385 300]{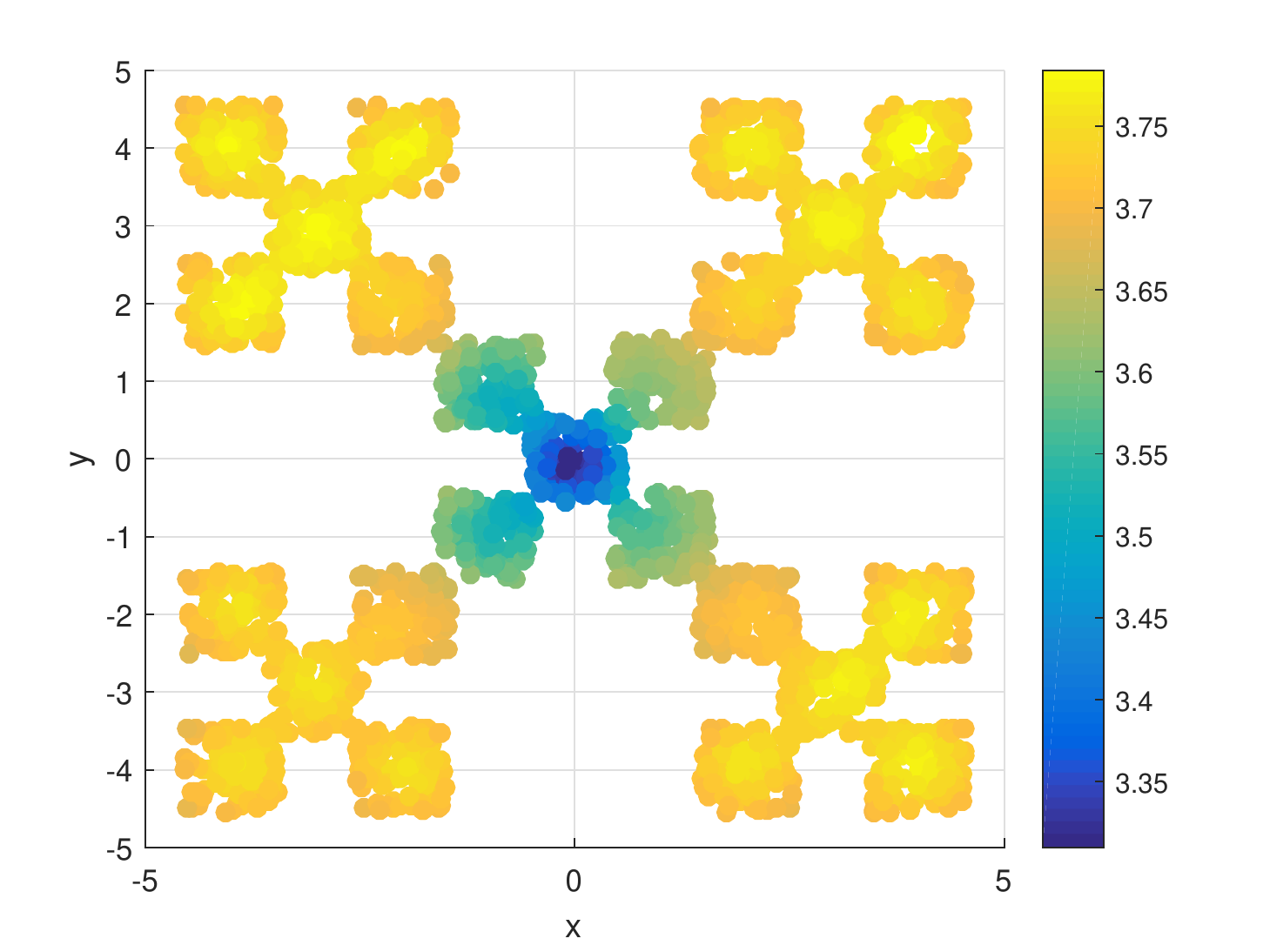}}
	\subfigure[$\hat{\mathbf{q}}_3$]{
		\includegraphics*[width=.23\columnwidth, viewport =48 30 385 300]{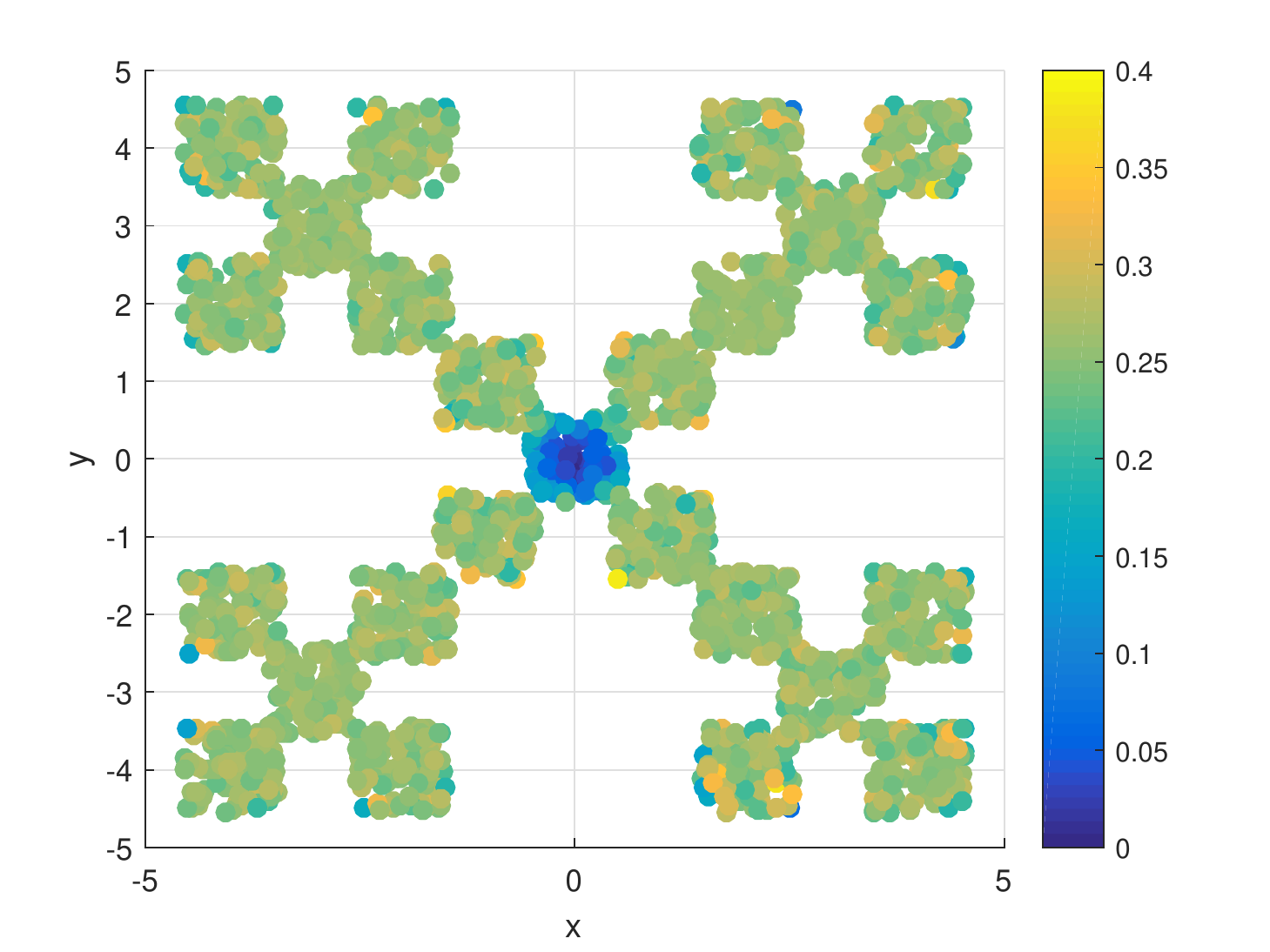}}
	\subfigure[$\hat{\mathbf{q}}_{8}$]{
		\includegraphics*[width=.23\columnwidth, viewport =48 30 385 300]{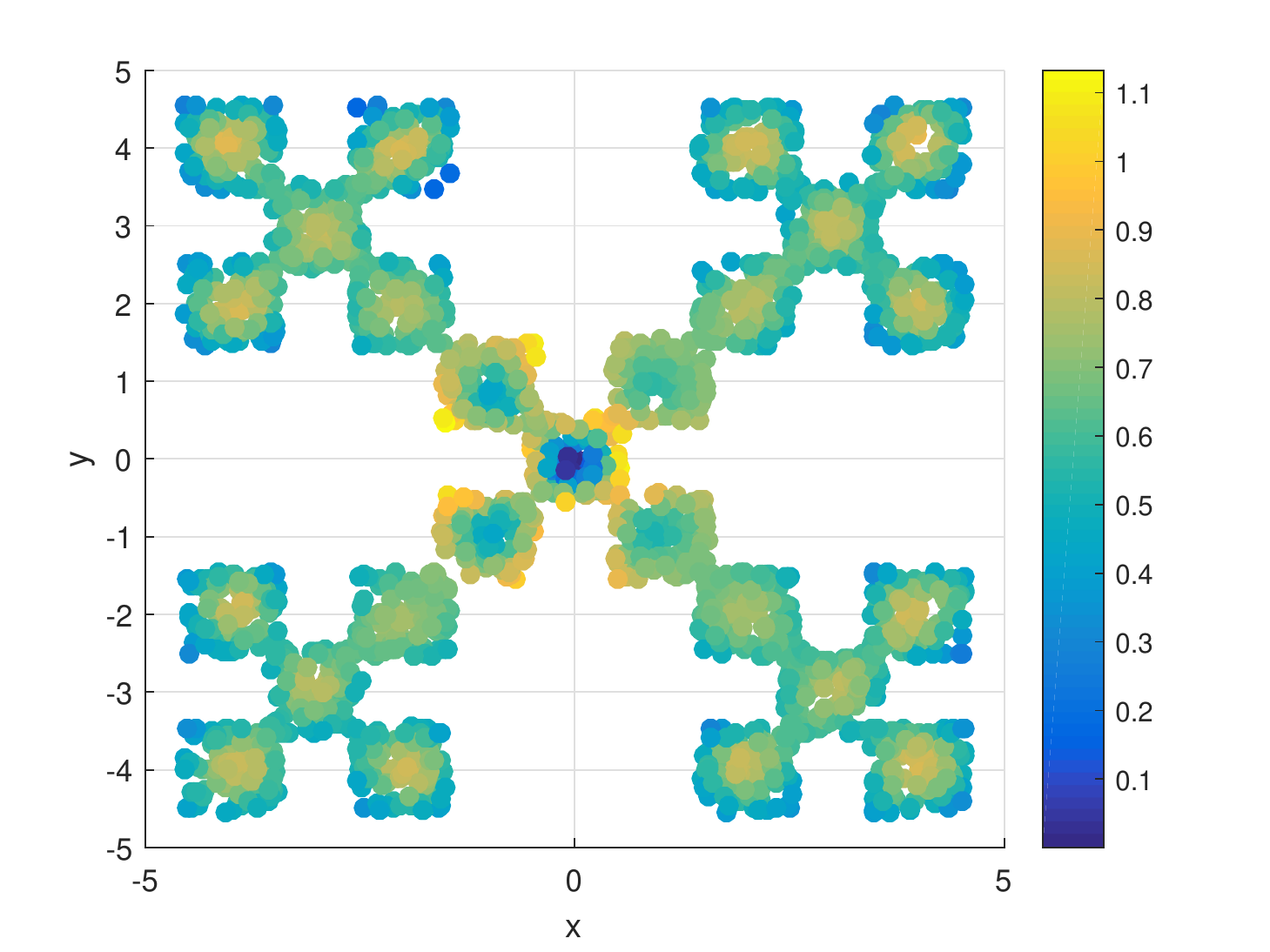}}
	\subfigure[]{
		\includegraphics*[width=.44\columnwidth]{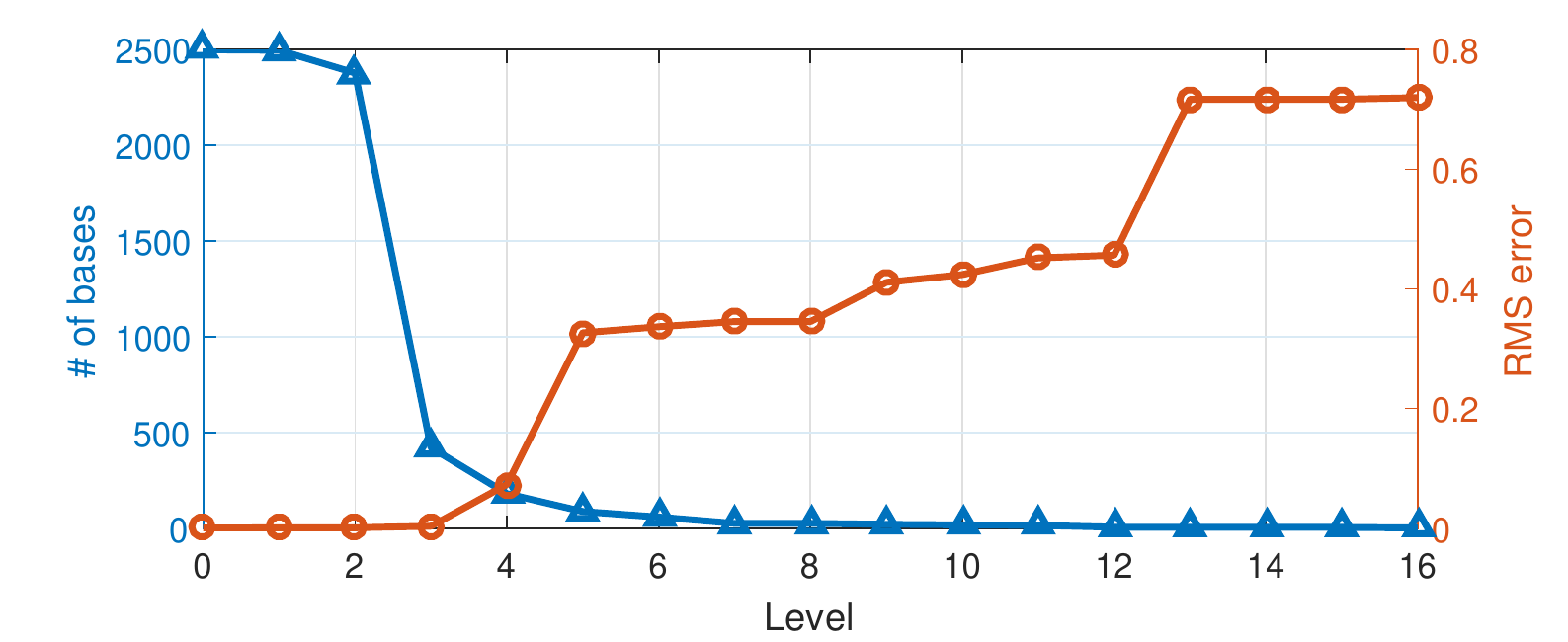}}
	\caption{Scaling functions at (a - b) level 3 and (c - d) level 8. (f - h) Exact and approximate value functions. (e, i - j) Exact and approximate cost functions. (k) The number of features, RMS error of the solution at each level.}
	\vspace*{-.15in}
	\label{fig:results}
\end{figure}
We consider a simple two-dimensional stochastic single integrator in the fractal-like environment.
The environment consists of 5 groups of rooms where one group is made of 5 square rooms as shown in Fig. \ref{fig:results};
one can observe that the environment has 2 level self-similarity, which makes the problem have a multiscale nature.
The dynamics is given by $\mathbf{f}(\mathbf{x}) = \mathbf{0},~G(\mathbf{x}) = I_2$; that is, the position of a robot in the configuration space, $\mathbf{x}\in \mathcal{X}$, is controlled by the velocity input, $\mathbf{u}\in\mathbb{R}^2$ while being disturbed by white noise.
We set $h=0.1$ and $\sigma = 1$.
In order to discretize the state space, 100 samples are obtained from each room, therefore there are 2500 discrete state in total.
The transition data set is obtained from true occupancy measure induced by the optimal policy, which is equivalent to using infinite number of samples.

Fig. \ref{fig:results} (a) - (d) shows multiscale features: some scaling functions in the diffusion wavelet tree at level 3 and 8.
It is seen that at level 3 and 8, where roughly $8h$ and $256h$ are considered as 1-step, scaling functions roughly represent each small room and one group of 5-rooms, respectively; it has no meaning to make a distinction within a room or a group of 5 rooms at those levels.
Fig. \ref{fig:results} (g, i) and (h, j) depict the approximated value and cost functions at level 3 and 8, respectively and Fig. \ref{fig:results} (k) shows the number of features and the RMS error of value functions at each scale.
At level 3, only 421 basis functions out of 3000 are used, but the value and cost functions are recovered quite exactly;
at level 8, where 25 bases are used, even though the solution contains some error, it interprets the information of optimal policy and the preference of cost function between the groups of 5 rooms.
We omit the computational time at each scale because of the space limitation but we observe that the computational time decreases as the number of feature increases;
It is obvious that there is a trade-off between the solution quality and the computational cost and the appropriate level can be chosen by observing the features at that level.

\section*{Acknowledgments}
This work was supported by Agency for Defense Development (under contract \#UD150047JD).
\bibliographystyle{plainnat}
\bibliography{nips16w}


\end{document}